\documentclass[runningheads]{llncs}

\usepackage[T1]{fontenc}
\usepackage{graphicx}
\usepackage{wrapfig}
\usepackage{booktabs}
\usepackage{multirow}
\usepackage{tabularx}
\usepackage{siunitx}

\let\llncssubparagraph\subparagraph
\let\subparagraph\paragraph
\usepackage{titlesec}
\titlespacing{\subsubsection}{0pt}{1.0ex plus 1ex minus .2ex}{1.0ex plus .2ex}
\let\subparagraph\llncssubparagraph

\usepackage[breaklinks,colorlinks,linkcolor=blue,urlcolor=blue,citecolor=blue]{hyperref} 

\usepackage{tikz}
\newcommand\copyrighttext{%
  \scriptsize This preprint has not undergone peer review or any post-submission improvements or corrections. The Version of Record of this contribution is published in \textit{Proceedings of the 29th Pacific-Asia Conference on Knowledge Discovery and Data Mining (PAKDD), Sydney, Australia, 10--13 June 2025}, and is available online at \href{https://doi.org/10.1007/978-981-96-8186-0_24}{https://doi.org/10.1007/978-981-96-8186-0\_24}.
  \smallskip
  
  \textcopyright\ 2025. Please cite this article as follows: G. Li, L. Chen, C. Tang, V. Švábenský, D. Deguchi, T. Yamashita, A. Shimada: \textit{Single-Agent vs. Multi-Agent LLM Strategies for Automated Student Reflection Assessment}. In Proceedings of the 29th Pacific-Asia Conference on Knowledge Discovery and Data Mining (PAKDD), Springer Nature, 2025. (Book: Advances in Knowledge Discovery and Data Mining. Chapter No: 24.) DOI: \href{https://doi.org/10.1007/978-981-96-8186-0_24}{10.1007/978-981-96-8186-0\_24}.}
\newcommand\copyrightnotice{%
\begin{tikzpicture}[remember picture,overlay]
\node[anchor=north,yshift=-24pt] at (current page.north) {\fbox{\parbox{\dimexpr\textwidth-\fboxsep-\fboxrule\relax}{\copyrighttext}}};
\end{tikzpicture}%
}

\begin{document}

\title{Single-Agent vs. Multi-Agent LLM Strategies for Automated Student Reflection Assessment}

\titlerunning{Single-Agent vs. Multi-Agent LLM Strategies for Reflection Assessment}

\author{
Gen Li\inst{1} \and
Li Chen\inst{1} \and
Cheng Tang\inst{1} \and
Valdemar Švábenský\inst{1}\orcidID{0000-0001-8546-280X} \and
Daisuke Deguchi\inst{2} \and
Takayoshi Yamashita\inst{3}\and
Atsushi Shimada\inst{1}
}

\authorrunning{G. Li et al.}

\institute{
Kyushu University, Fukuoka, Japan\\
\email{\{gen.li, chenli, tang, atsushi\}@limu.ait.kyushu-u.ac.jp},
\email{valdemar@kyudai.jp} \and
Nagoya University, Nagoya, Japan\\
\email{ddeguchi@nagoya-u.jp} \and
Chubu University, Kasugai, Japan\\
\email{takayoshi@isc.chubu.ac.jp}
}

\maketitle

\begin{abstract}
We explore the use of Large Language Models (LLMs) for automated assessment of open-text student reflections and prediction of academic performance. Traditional methods for evaluating reflections are time-consuming and may not scale effectively in educational settings. In this work, we employ LLMs to transform student reflections into quantitative scores using two assessment strategies (single-agent and multi-agent) and two prompting techniques (zero-shot and few-shot). Our experiments, conducted on a dataset of 5,278 reflections from 377 students over three academic terms, demonstrate that the single-agent with few-shot strategy achieves the highest match rate with human evaluations. Furthermore, models utilizing LLM-assessed reflection scores outperform baselines in both at-risk student identification and grade prediction tasks. These findings suggest that LLMs can effectively automate reflection assessment, reduce educators' workload, and enable timely support for students who may need additional assistance. Our work emphasizes the potential of integrating advanced generative AI technologies into educational practices to enhance student engagement and academic success.


\copyrightnotice
\end{abstract}

\section{Introduction}

\begin{wrapfigure}{r}{0.5\columnwidth}
    \centering
    \includegraphics[width=\linewidth]{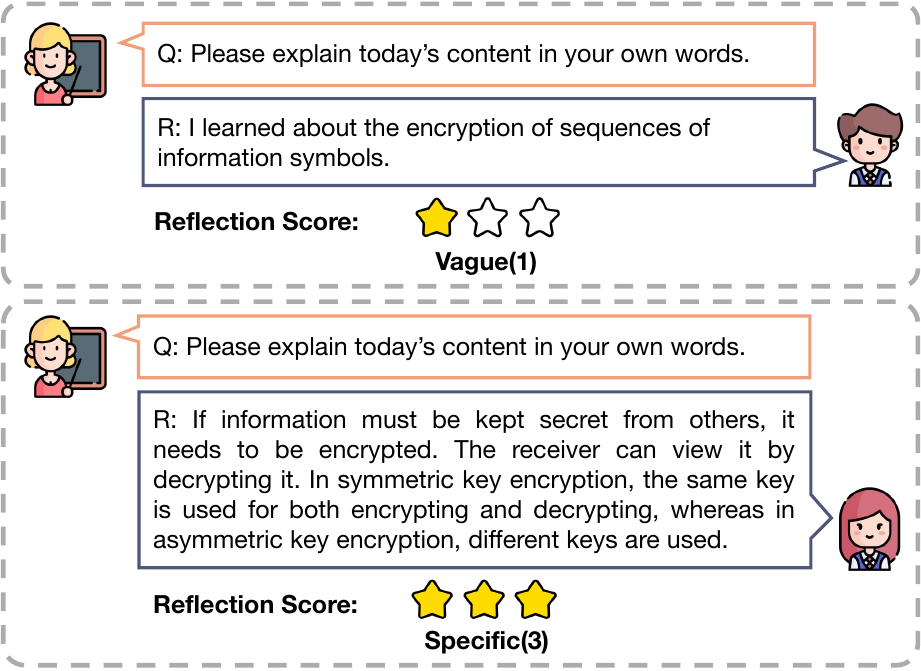}
    \caption{Examples of student reflections and the corresponding assessed scores.}
    \label{ref_intro}
\end{wrapfigure}

In today's educational environments, learners generate a substantial amount of open-ended textual data, such as essays, discussion posts, and reflections. Among these, student reflections are particularly valuable as they offer deep insights into learners' understanding and experiences. As illustrated in Fig \ref{ref_intro}, reflections are typically prompted by educators to encourage students to think back on and articulate their learning after engaging with new material or completing activities \cite{Ullmann2019Automated}. These reflective exercises not only aid students in self-assessing their comprehension \cite{zimmerman2002becoming} but also provide educators with a window into students' cognitive processes and how they integrate new knowledge \cite{reflectCore}. Engaging in reflective practice has been shown to positively impact learning outcomes \cite{Chang2011EFFECTS,steiner2016strategy}.

Despite the richness of information contained in student reflections, systematically analyzing these unstructured responses poses significant challenges. The open-text nature of reflections leads to variability in content and expression styles, making manual assessment both time-consuming and complex. Early efforts focused on developing rubrics for manually evaluating reflection levels from various perspectives \cite{ash,sabourin2013understanding,menekse}. More recent studies explored automated methods using machine learning and natural language processing to assess reflection quality \cite{Ullmann2019Automated,reflectionquality}.

Advancements in LLMs, such as ChatGPT, offer new possibilities for addressing these challenges. LLMs have demonstrated exceptional capabilities in understanding and processing complex textual data, including the ability to follow detailed instructions and apply evaluation criteria consistently \cite{zheng2023judging}. This indicates potential for leveraging LLMs to automate the assessment of reflections by transforming qualitative responses into quantitative scores based on specified criteria. Moreover, using prompting strategies, such as zero-shot and few-shot learning, can guide LLMs' assessment processes effectively \cite{wang2020generalizing}.

Recent studies have investigated the use of LLMs for coding open-text data, showing substantial alignment with expert evaluations \cite{lakgpt,chenthree}. However, performing LLM-assisted automatic assessment of student reflections, and its use to predict academic performance, remains underexplored. Additionally, verifying the consistency and reliability of LLM-generated assessments compared to human evaluations is essential to ensure their effectiveness in educational settings.

To address these gaps, this study explores the use of LLMs to quantitatively assess student reflections and predict academic performance, including both at-risk identification and grade prediction. We employ different prompting strategies, including single-agent and multi-agent configurations combined with zero-shot and few-shot learning, to guide the LLM's assessment process. Furthermore, we incorporate human labels to verify the consistency of the LLM's assessments across different prompting methods.

We collected data from real educational contexts at Kyushu University over three academic terms, involving 377 students and 5,278 reflections. We evaluate the consistency of LLM's assessed reflection scores by comparing them with human labels across different prompting strategies. Additionally, we assess the effectiveness of these scores in predicting academic performance, including at-risk identification and grade prediction, using various machine learning models. Our approach offers a scalable and efficient solution for analyzing reflective writings, demonstrating its potential in enhancing educational analytics and student support mechanisms. The main contributions of this paper are as follows:

\begin{enumerate}
    \item \textbf{LLM-Assisted Reflection Assessment}: We propose a novel automated method using LLMs to assess student reflections quantitatively, converting open-text responses into numerical scores that reflect levels of understanding and engagement.
    \item \textbf{Prompting Strategies and Human Validation}: We experiment with different prompting strategies (i.e., single-agent and multi-agent configurations combined with zero-shot and few-shot learning), and validate the consistency of the assessments by comparing them with human labels.
    \item \textbf{Empirical Evaluation and Academic Performance Prediction}: We evaluate our method using real educational data, demonstrating its effectiveness in enhancing educational analytics and improving student support through the identification of at-risk students and accurate grade prediction.
\end{enumerate}

\section{Related Work}

\subsection{Student Reflection}
Reflective practice is an important component in educational contexts~\cite{Ukrop2019}, enabling students to assess their progress, understand their learning processes, and adjust strategies accordingly~\cite{zimmerman2002becoming}. Studies have highlighted the influence of reflective practice on academic achievement~\cite{Yan2020reflection}, and have categorized reflections based on their focus, such as specific reflections on learning activities and general reflections on overall progress~\cite{sabourin2013understanding}.

Interventions designed to enhance student reflection have shown positive effects on academic performance~\cite{long2013supporting,steiner2016strategy}. Additionally, active cognitive strategies, like summarizing and creating analogies, have been found to enhance the reflection process and contribute to better learning results~\cite{weinstein2011self,broadbent2017comparing}.

The assessment of reflection quality has attracted attention, with researchers developing rubrics to evaluate reflections and exploring automated methods for assessment. Rubrics defining key dimensions of critical reflection have been proposed~\cite{ash}, and studies have shown positive correlations between reflection quality and learning gains~\cite{menekse}. Efforts to automate reflection assessment using features derived from such rubrics offer scalable alternatives to manual evaluations~\cite{reflectionquality}. However, prior work has not yet attempted to systematically evaluate the assessment of reflections using LLMs. Our work aims to address this gap.

\subsection{Academic Performance Prediction}

Predicting student academic performance has been a significant focus in educational data mining research, utilizing various data sources and methodologies to enable early interventions and support student success. Early studies integrated data such as grades, demographic characteristics, and learning management system (LMS) interactions to predict performance and assign risk levels~\cite{arnold2012course}. Subsequent research expanded on this by incorporating attendance patterns~\cite{EDM-posters74}, course engagement metrics~\cite{conijn}, course types~\cite{Brown2016WhatAW}, and a lecture quiz~\cite{li2024llm}.

Combining institutional data with LMS data has been shown to improve prediction accuracy more than using either data source alone~\cite{Yu2020TowardsAA}. Holistic frameworks that integrate psychological, cognitive, economic, and institutional variables emphasize the importance of a multidimensional approach to accurately predict and support student performance~\cite{Adejo}.

Recent research has also addressed fairness and equity in predictive models. Advanced techniques such as adversarial learning and equity-based sampling have been employed to reduce biases in grade prediction algorithms, aiming to support historically underserved student groups effectively~\cite{jiang2021towards}.

Our work builds upon these studies by leveraging LLMs to assess student reflections and predict academic performance, including both at-risk identification and grade prediction. By transforming qualitative reflection data into quantitative scores, we offer a novel approach that uses data about students' self-reported learning experiences to perform predictive modeling.

\section{Methods}

\subsection{Problem Statement}

In this study, we automate the assessment of student reflections and predict academic performance using the assessed scores. Our goal is to leverage LLMs to transform qualitative, open-ended reflections into quantitative scores. Subsequently, we can utilize these scores to predict whether students are at risk of underperforming academically. We formally define the problem as follows:

\textbf{Student Reflections:} Consider a set of students $\mathcal{S} = \{s_1, s_2, \dots, s_N\}$ enrolled in a course consisting of $T$ sessions (e.g., weeks). After each session $t$ ($1 \leq t \leq T$), students are prompted with a set of reflective questions $\mathcal{Q} = \{q_1, q_2, \dots, q_M\}$. Each student $s_i$ responds to these questions, resulting in a collection of reflections $\mathcal{R}_{i} = \{ r_{i,j,t} \mid q_j \in \mathcal{Q},\, 1 \leq t \leq T \}$,
where $r_{i,j,t}$ denotes the reflection of student $s_i$ to question $q_j$ at session $t$.

\textbf{LLM-Based Reflection Assessment:} Our first objective is to employ an LLM to assess each reflection $r_{i,j,t}$ and assign a score $s_{i,j,t} \in \{0, 1, 2, 3\}$, representing the quality of the reflection according to predefined criteria. The assessment process can be formulated as $s_{i,j,t} = $ LLM$(r_{i,j,t}, P)$, where $P$ represents the prompting strategy and assessment criteria provided to the LLM.

\textbf{Academic Performance Prediction:} Our second objective is to utilize the assessed scores to predict students' academic performance. Let $\mathbf{S}_i$ be the set of all scores for student: $\mathbf{S}_i = \{ s_{i,j,t} \mid q_j \in \mathcal{Q},\, 1 \leq t \leq T \}$, We aim to predict the final grade $g_i$ of student $s_i$ based on their reflection scores $\mathbf{S}_i$. Specifically, we seek to learn a predictive function $f$ such that $\hat{g}_i = f(\mathbf{S}_i)$, where $\hat{g}_i$ is the predicted grade or risk status (e.g., at-risk or not at-risk).

\subsection{Reflection Assessment}

This section details our approach to assessing student reflections, including the criteria used and the implementation of LLMs for automated scoring.

\subsubsection{Assessment Criteria:}

Effective reflection assessment requires robust criteria to ensure accurate and meaningful evaluation. We employ the well-established four-level scoring system in the education field from \cite{menekse}, which consists of the \textbf{Scoring Criteria} and the corresponding \textbf{Decision Tree Rubric}.

\begin{figure*}[htbp]
\vspace{-20pt}
\centering 
\includegraphics[width=\textwidth]{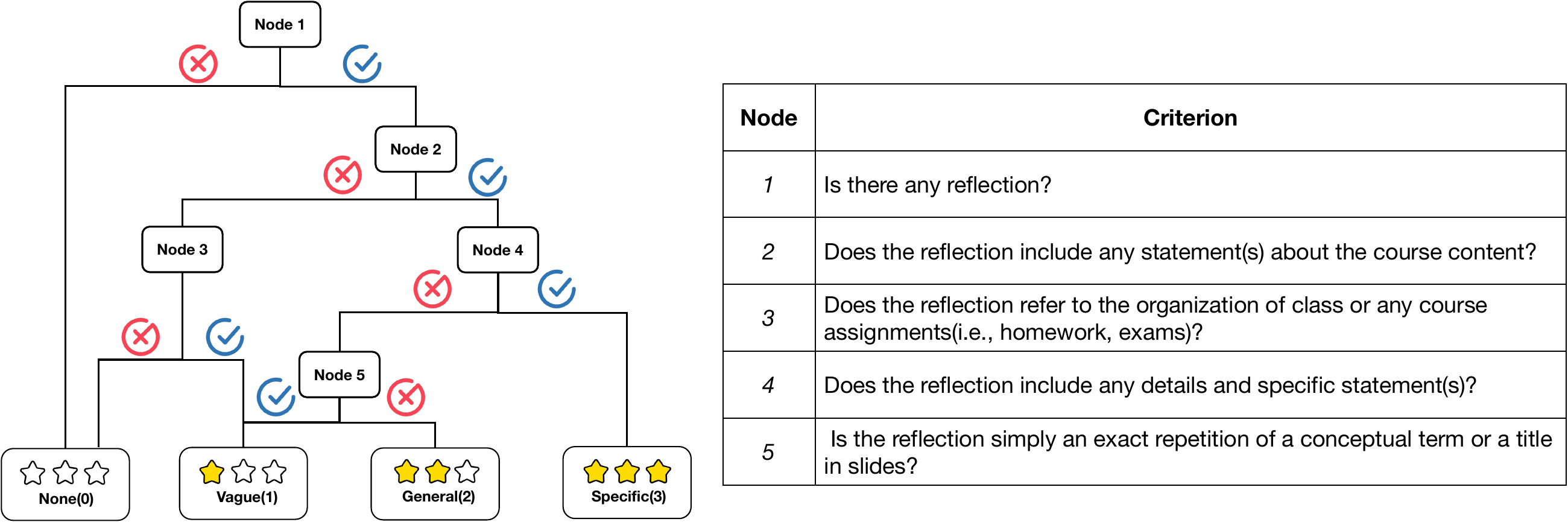}  
\caption{The Decision Tree Rubric for reflection assessment. Each node represents a criterion that guides the evaluator to the appropriate score.}  
\label{Rubric}   
\end{figure*}

The \textbf{Scoring Criteria} provide general descriptions for each score level, as detailed in Table \ref{tab:criteria-assessment}. The \textbf{Decision Tree Rubric}, illustrated in Figure \ref{Rubric}, outlines a step-by-step process where each node corresponds to a specific question or criterion that the evaluator verifies to determine the appropriate score.

\begin{table*}[htbp]
\caption{The four-level Scoring Criteria for reflection assessment.}
\centering
\begin{tabularx}{\columnwidth}{>{\hsize=.3\hsize}X|>{\hsize=1.7\hsize}X}
\toprule
\textbf{Score} & \textbf{Scoring Criteria} \\
\hline
3 (Specific) & The reflection is specific and highly detailed, demonstrating deep understanding and engagement. \\
\hline
2 (General)  & The reflection goes beyond broad concept statements but lacks depth or specific details. \\
\hline
1 (Vague)    & The reflection contains only broad concepts with little or no explanation. \\
\hline
0 (None)     & No reflection provided, or the reflection is irrelevant or unrelated to the course content. \\
\bottomrule
\end{tabularx}
\label{tab:criteria-assessment}
\end{table*}

\subsubsection{LLM-Based Assessment:}

\begin{figure*}[ht]
    \centering 
    \includegraphics[width=\textwidth]{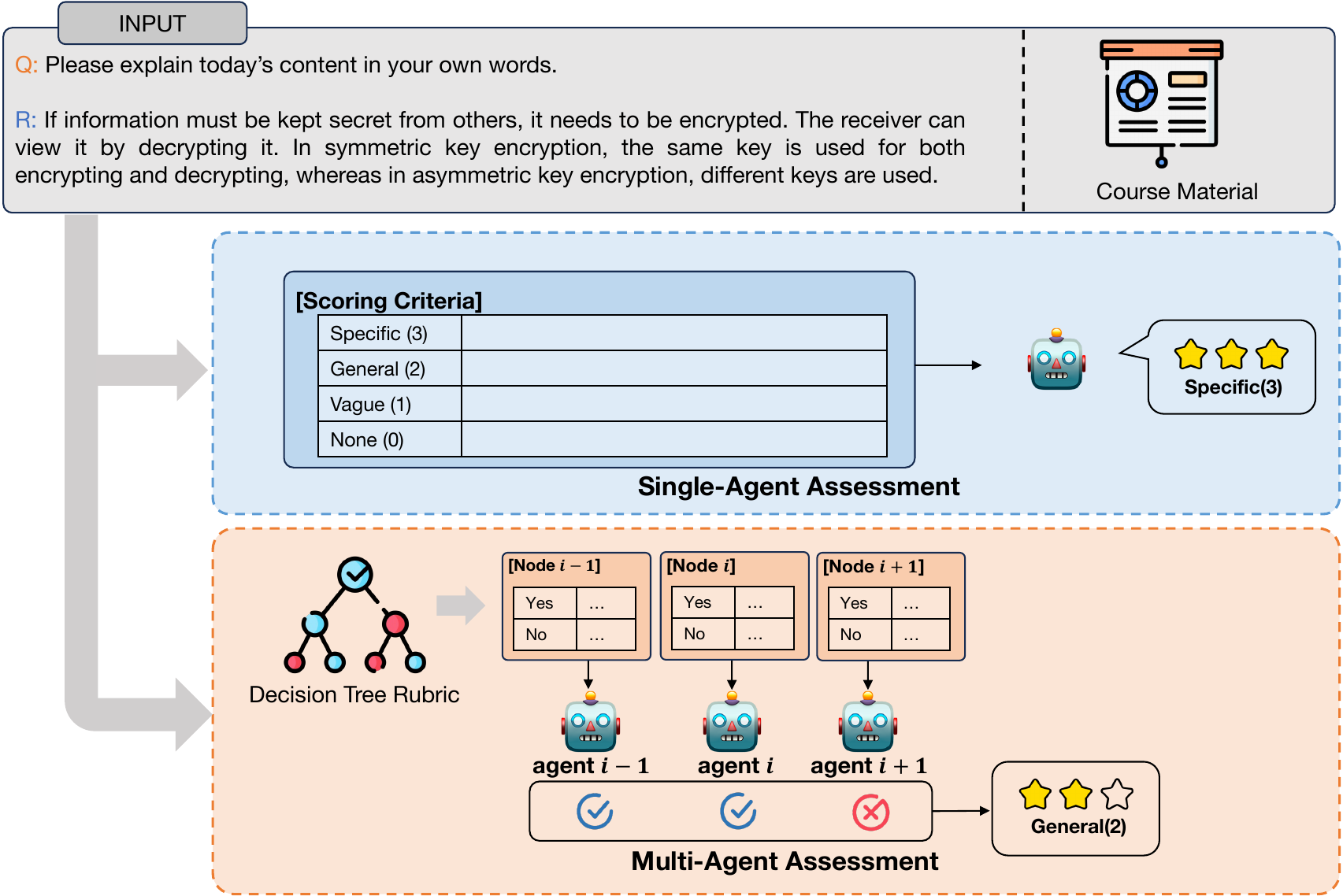}  
    \caption{Instruction strategies for LLMs to assess student reflections: (upper) Single-Agent Assessment using the Scoring Criteria and (lower) Multi-Agent Assessment using the Decision Tree Rubric.}
    \label{instruction}   
\end{figure*}

Leveraging the capabilities of LLMs, we implemented two distinct assessment strategies that mirror the human evaluation methods: single-agent assessment and multi-agent assessment. Within each strategy, we explored different prompting approaches (i.e., zero-shot and few-shot prompting) to guide the LLM's evaluation process, as illustrated in Fig. \ref{instruction}.

\paragraph{Single-Agent Assessment using the Scoring Criteria:}

In this approach, a single LLM acts as the evaluator. The LLM is given the \textbf{Scoring Criteria} descriptions for each score level (as detailed in Table \ref{tab:criteria-assessment}), along with the student's reflection $r_{i,j,t}$. The LLM then assigns a score $s_{i,j,t}$ based on how the reflection meets with the criteria. We employed two prompting techniques in this strategy:

\begin{itemize}
    \item Zero-Shot Prompting: The LLM is given the scoring criteria and the reflection without any example assessments. It relies solely on the provided criteria to evaluate the reflection.

    \item Few-Shot Prompting: In addition to the scoring criteria, the LLM is provided with a few example reflections and their corresponding scores (one example for each score level).
\end{itemize}

\paragraph{Multi-Agent Assessment using the Decision Tree Rubric:}

This strategy involves multiple LLM agents collaborating to assess the reflection, implementing the step-by-step process of the \textbf{Decision Tree Rubric} (Figure \ref{Rubric}). Each agent is responsible for evaluating a specific criterion or question in the decision tree. Starting from the root, agents sequentially determine the answers (Yes/No) to the criteria at each node based on the reflection $r_{i,j,t}$. The collective decisions of the agents reach the final score $s_{i,j,t}$ according to the tree. This method mimics human evaluators who systematically follow the rubric for assessment. Furthermore, we applied zero-shot and few-shot prompting in the assessment:

\begin{itemize}
    \item Zero-Shot Prompting: Each agent evaluates its assigned criterion without any example assessments, only relying on the criterion description.

    \item Few-Shot Prompting: Agents are provided with example reflections and the corresponding decisions (Yes/No) at their respective nodes. This helps the agents understand how to apply the criteria based on examples.
\end{itemize}

\section{Experiments}
\subsection{Dataset}

We conducted our study within the real educational setting of Kyushu University, specifically in the \textit{Information Science} course. Reflective practice was integrated into the curriculum over three academic terms, each enrolling different sets of students. Each term spanned 14 weeks, concluding with a final examination. Each week, following the lecture, students were asked to respond to reflective questions designed to capture their learning experiences and comprehension:

\begin{quote}
\textbf{\textit{Reflect on today's lesson by explaining the main concepts in your own words, describing what you understood and can now apply, and sharing any additional thoughts or insights you gained.}} (Translated from the original language, i.e., Japanese.)
\end{quote}

Therefore, each student provided a total of 14 open-text responses from the 14 weeks. We collected reflections from 377 students across the three terms, resulting in a dataset of 5,278 reflections. Additionally, we collected the final grades for each student, classified into categories A--E. There were 219 students (58\%) with grades A and B categorized as \textit{no-risk}, while the remaining 158 students (42\%) with grades C, D, and E were classified as \textit{at-risk}. Table \ref{tab:grades} shows the distribution of student grades across the terms.

\setlength{\intextsep}{0pt}
\begin{wraptable}{r}{0.5\linewidth}
\setlength{\tabcolsep}{5pt}
\caption{Distribution of student final grades in each term.}
\smallskip
\centering
\scriptsize
\begin{tabular}{l|rrrrr|r}
\toprule
\textbf{Term} & \textbf{A} & \textbf{B} & \textbf{C} & \textbf{D} & \textbf{E} & \textbf{Total} \\ 
\midrule
Term 1 & 9  & 53 & 32 & 7  & 6  & 107 \\
Term 2 & 15 & 88 & 37 & 9  & 25 & 174 \\
Term 3 & 17 & 37 & 34 & 4  & 4  & 96  \\ 
\midrule
Total  & 41 & 178& 103& 20 & 35 & 377 \\ 
\bottomrule
\end{tabular}
\label{tab:grades}
\end{wraptable}

The data collection adhered to ethical guidelines of Kyushu University to ensure the privacy and confidentiality of all participants. Informed consent was obtained from all students prior to the study. Participants were assured that their reflections and grades would be anonymized and used solely for research purposes.

\subsection{Reflection Assessment and Validation with Human Labels}

We utilized GPT-4o with multi-agent framework~\cite{liang2023encouraging} to conduct the quantitative assessment of student reflections. As shown in Table \ref{tab:criteria-assessment}, the scoring rubric ranges from 0 to 3, and we employed two assessment strategies $\times$ two prompting techniques, resulting in four combinations of conditions: \textit{single-agent zero-shot}, \textit{single-agent few-shot}, \textit{multi-agent zero-shot}, and \textit{multi-agent few-shot}.

Furthermore, to evaluate the consistency of LLM's assessments, we manually scored all student reflections from Term 1 following the Scoring Criteria and Decision Tree Rubric. There were three qualified evaluators: two research assistants and one professor with expertise in the domain. To ensure reliability, we calculated the inter-rater agreement using Krippendorff's Alpha, confirming a score of 0.8386, which indicates strong agreement among the evaluators. We then compared the LLM-generated scores with the human-labeled scores. Exact Match (EM) rate is used to evaluate LLM assessment performance against human labels, which measures the percentage of cases where LLM scores exactly match human-labeled scores.

\subsection{Predictive Modeling}

We utilized the assessed reflection scores \( s_{i,j,t} \) to predict students' academic performance, focusing on two tasks:

\begin{enumerate}
    \item \textbf{At-Risk Identification}: Predicting whether a student is at-risk (grades C, D, E) or not at-risk (grades A, B).
    \item \textbf{Grade Prediction}: Predicting the specific final grade category (A--E) for each student.
\end{enumerate}

We used data from Terms 1 and 2 to train the models and data from Term 3 as a holdout test set. This setup simulates a realistic deployment scenario where models are built on historical data and evaluated on a new cohort of students, assessing model generalizability. We employed three machine learning models:

\begin{itemize}
    \item \textbf{Recurrent Neural Networks (RNNs)}: Specifically, we used Long Short-Term Memory (LSTM) networks to handle the sequential reflection scores.
    \item \textbf{Random Forest (RF)}: A traditional classifier that uses aggregated features from the reflection scores.
    \item \textbf{BERT-based Classifier} (baseline): Uses the raw text of student reflections as input without quantitative scoring.
\end{itemize}

We evaluated the models using common classification metrics: Accuracy, Precision, Recall, and F1-Score.

\subsection{Results and Discussion}

\subsubsection{Consistency with Human Labels:}

\setlength{\intextsep}{0pt}
\begin{wraptable}{r}{0.4\linewidth}
\vspace{-10pt}
\caption{EM Rates (\%) of LLM Assessments vs. Human Labels.}
\smallskip
\centering
\scriptsize
\begin{tabular}{l|cc}
\toprule
\textbf{Agent Type} & \textbf{Zero-Shot} & \textbf{Few-Shot} \\
\midrule
Single-Agent & 74.9$\pm$3.96 & \textbf{82.8$\pm$3.63} \\
Multi-Agent  & 75.7$\pm$2.74 & 79.5$\pm$2.77 \\
\bottomrule
\end{tabular}
\label{tab:consistency_rate}
\end{wraptable}

\begin{figure}[htbp]
    \centering 
    \includegraphics[width=.85\linewidth]{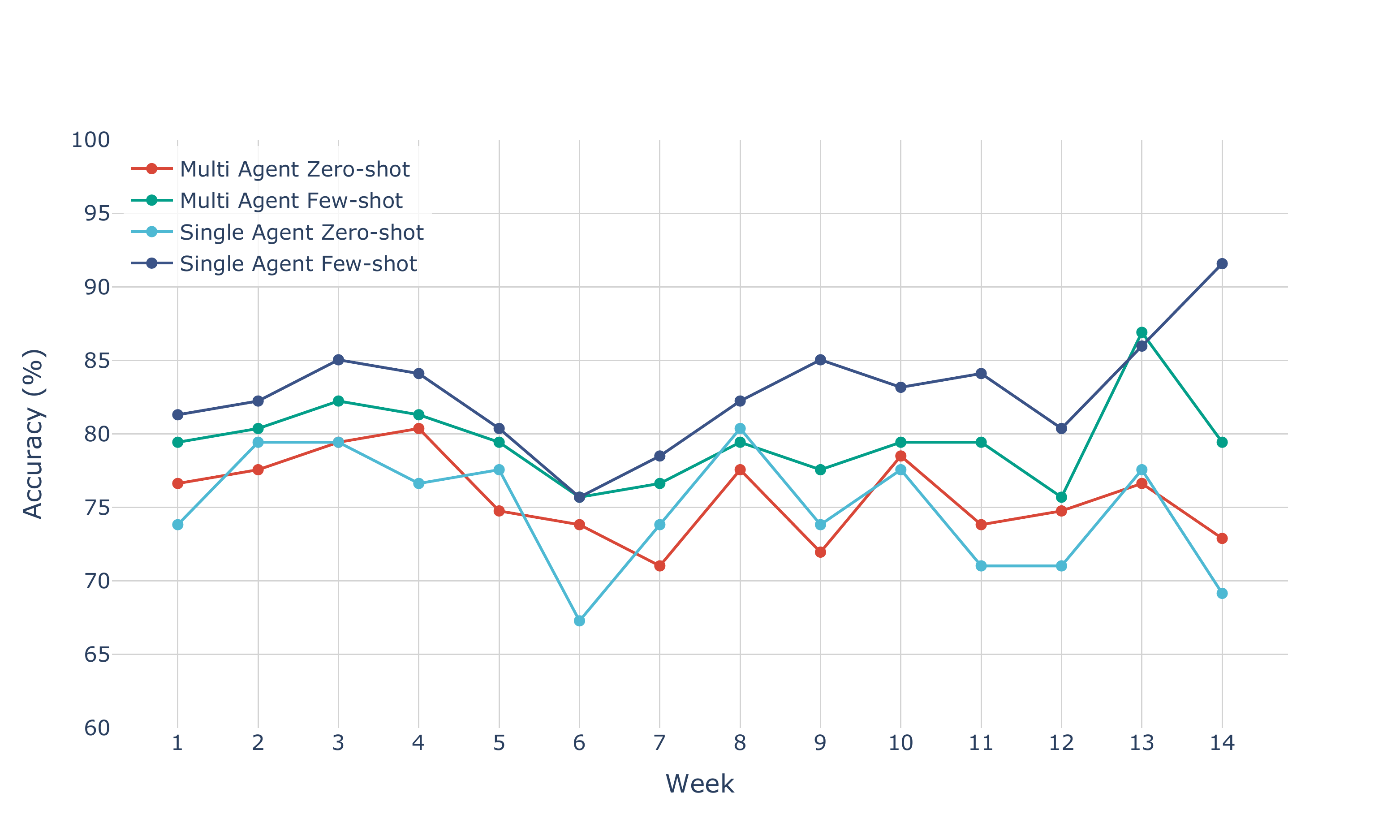}  
    \caption{EM rates across weeks for different agent and prompting settings.}
    \label{fig:accuracy_trends}   
\end{figure}

Table \ref{tab:consistency_rate} presents the EM rates of LLM's assessments for Term 1 reflections, while Figure \ref{fig:accuracy_trends} illustrates the EM rate trends across the 14 weeks for different agent and prompting settings. Providing few-shot examples clearly improves the rate, particularly for the single-agent approach, where the EM rate increases from 74.9\% to 82.8\%. The multi-agent approach also benefits from few-shot prompting, with EM rates increasing from 75.7\% to 79.5\%. Moreover, the multi-agent method is more stable, as indicated by the lower standard deviations (2.74 and 2.77) compared to the single-agent method (3.96 and 3.63), suggesting that the multi-agent approach performs more consistently across different weeks.

\subsubsection{Academic Performance Prediction:}

Table \ref{tab:model_performance} presents the performance metrics for at-risk identification and grade prediction. The models using LLM-assessed reflection scores outperform the baseline model that uses raw text inputs. The highest accuracy was achieved using the single-agent few-shot input for both at-risk identification and grade prediction. Specifically, for at-risk identification, the LSTM model achieved an accuracy of 79.3\% using single-agent few-shot assessments. For grade prediction, the Random Forest model achieved an accuracy of 55.5\% using the same input. 

\bigskip

\begin{table*}[htbp]
\setlength{\tabcolsep}{2.5pt}
\caption{Performance comparison of baseline and our models with different inputs.}
\centering
\scriptsize
\begin{tabular}{lll*{8}{S[table-format=2.1]}}
\toprule
\multirow{2}{*}{\textbf{Approach}} & \multirow{2}{*}{\textbf{Model}} & \multirow{2}{*}{\textbf{Input}} & \multicolumn{4}{c}{\textbf{At-Risk Prediction}} & \multicolumn{4}{c}{\textbf{Grade Prediction}} \\
\cmidrule(lr){4-7} \cmidrule(lr){8-11}
 &  &  & {Acc} & {Prec.} & {Recall} & {F1} & {Acc} & {Prec.} & {Recall} & {F1} \\
\midrule
\textbf{Baseline} & BERT & Raw Text & 69.8 & 74.1 & 69.8 & 66.7 & 46.9 & 50.3 & 46.9 & 42.4 \\
\midrule
\multirow{8}{*}{\textbf{Ours}} & \multirow{4}{*}{LSTM}
   & Single-Agent Zero-Shot & 75.9 & 77.5 & 75.9 & 74.8 & 55.3 & 47.4 & 55.3 & 49.6 \\
 &  & Single-Agent Few-Shot & \textbf{79.3} & \textbf{81.1} & \textbf{79.3} & \textbf{78.5} & 54.8 & 46.6 & 54.8 & 48.6 \\
 &  & Multi-Agent Zero-Shot & 76.7 & 78.5 & 76.7 & 75.6 & 53.6 & 47.0 & 53.6 & 47.7 \\
 &  & Multi-Agent Few-Shot & 75.0 & 76.2 & 75.0 & 74.1 & 53.5 & 47.9 & 53.5 & 47.2 \\
\cmidrule(lr){2-11}
 & \multirow{4}{*}{RF}
   & Single-Agent Zero-Shot & 73.5 & 73.7 & 73.5 & 73.0 & 48.0 & 41.9 & 48.0 & 42.9 \\
 &  & Single-Agent Few-Shot & 76.6 & 78.6 & 76.6 & 75.5 & \textbf{55.5} & \textbf{56.8} & \textbf{55.5} & \textbf{51.7} \\
 &  & Multi-Agent Zero-Shot & 76.4 & 78.0 & 76.4 & 75.5 & 49.0 & 41.9 & 49.0 & 42.0 \\
 &  & Multi-Agent Few-Shot & 76.7 & 77.9 & 76.7 & 75.9 & 53.0 & 44.5 & 53.0 & 45.3 \\
\bottomrule
\end{tabular}
\label{tab:model_performance}
\end{table*}

\subsubsection{Discussion of Instruction Strategies:}

Our results reveal the effectiveness of different instruction strategies. The differences are particularly notable when comparing the performance between single-agent and multi-agent approaches under zero-shot and few-shot prompting.

\paragraph{Single-Agent vs. Multi-Agent:}

Comparing the single-agent and multi-agent approaches, we observe that the multi-agent strategy performs better in the zero-shot setting, whereas the single-agent strategy excels in the few-shot setting. This pattern suggests that the multi-agent approach is more robust when examples are not available, possibly due to its structured and explicit decision-making process that provides clear guidance at each assessment step.

However, when examples are provided, the single-agent approach outperforms the multi-agent approach. This may be because the single-agent LLM can utilize in-context learning more effectively, incorporating the examples into its holistic assessment. The multi-agent approach, being distributed across multiple agents, may not benefit as much from the examples due to the complexity of coordinating and integrating information across agents, potentially leading to diminished gains from few-shot prompting.

\paragraph{Implications:}

These findings suggest that the choice of assessment strategy should be based on the availability of example reflections and the desired balance between stability and adaptability. If no examples are available, the multi-agent approach may provide better guidance to the LLM due to its structured framework. However, if examples can be provided, the single-agent approach is likely to yield better performance by effectively leveraging in-context learning to enhance assessment accuracy.

In practical applications, providing example reflections may require additional effort but can significantly improve the effectiveness of the single-agent approach. Educators and researchers should consider the trade-offs between the ease of implementation and the potential gains in performance.

\section{Conclusion}

This study explored the use of LLMs for the automated assessment of student reflections and the prediction of academic performance. We employed two assessment strategies (single-agent and multi-agent) and two prompting techniques (zero-shot and few-shot). Our experiments confirmed high EM rate with human evaluations and showed strong performance in key predictive applications of educational data mining. The findings suggest that LLMs like GPT-4o can effectively automate the assessment of student reflections, reducing the workload of educators and enabling timely identification of students who may need additional support.

Future work may extend the study beyond a single course to diverse educational contexts, explore the use of different LLMs, and address potential biases in LLM assessments. By further refining these methods, we can enhance the integration of AI technologies in education to support student success.

\section*{Acknowledgments}
This work was supported by JST CREST Grant Number JPMJCR22D1 and JSPS KAKENHI Grant Number JP22H00551, Japan.

\bibliographystyle{splncs04}
\bibliography{reference}

\end{document}